\documentclass[runningheads]{llncs}
\usepackage[T1]{fontenc}
\usepackage{graphicx,verbatim}
\usepackage{amsmath,amssymb}
\usepackage{booktabs}
\usepackage{subcaption}
\usepackage{algorithm}
\usepackage{algpseudocode}
\usepackage{color}
\usepackage[breaklinks,colorlinks=false]{hyperref}

\newenvironment{acknowledgments}{\section*{Acknowledgements}}{}
\begin{document}

\title{Skin Lesion Classification Based on ResNet-50 Enhanced With Adaptive Spatial Feature Fusion}
\titlerunning{Skin Lesion Classification via Adaptive Spatial Feature Fusion in ResNet-50}

\author{
Runhao Liu\inst{1}\textsuperscript{*,\dag}
\and
Fengyi Zha\inst{2}\textsuperscript{*}
\and
Fei Ding\inst{3}\textsuperscript{*}
\and\\
Guangzhen Yao\inst{4}
\and
Peng Zhang\inst{5}\textsuperscript{\dag}
}
\authorrunning{R. Liu et al.}

\institute{
$^1$Polytechnic Institute, Zhejiang University, Hangzhou, China\\
\email{runhaoliu@zju.edu.cn}\\
$^2$Chu Kochen Honors College, Zhejiang University, Hangzhou, China\\
$^3$Alibaba Group, Chaoyang District, Beijing, China\\
$^4$School of Information Science and Technology, Northeast Normal University, Changchun, China\\
$^5$School of Mathematical Sciences, Zhejiang University, Hangzhou, China\\
\email{pengz@zju.edu.cn}
}
  
\maketitle              

\renewcommand{\thefootnote}{}
\footnotetext{* These authors contributed equally.\\
\dag~Corresponding Author}
\renewcommand{\thefootnote}{\arabic{footnote}}
\begin{abstract}
Skin cancer classification is challenging due to high inter-class similarity, intra-class variability, and artifacts in dermoscopic images. To address these issues, we propose an improved ResNet-50 with Adaptive Spatial Feature Fusion (ASFF), which adaptively integrates multi-scale semantic and surface features to refine representations and reduce overfitting. The ResNet-50 model is enhanced with an adaptive feature fusion mechanism to achieve more effective multi-scale feature extraction and improve overall performance. Specifically, a dual-branch design fuses high-level semantic and mid-level detail features which use global average pooling and fully connected layers to produce spatial weights, and emphasizes lesion-relevant regions. Evaluated on a balanced subset of \textbf{ISIC 2020} (3,297 images, randomly selected from the original dataset), the ASFF-based ResNet-50 outperforms multiple CNN baselines, achieving \textbf{93.182\%} accuracy with superior precision, recall, specificity, and F1. It also reaches 0.9670 AUC (P-R) and 0.9717 AUC (ROC). \textbf{Grad-CAM} visualizations show more accurate focus on lesion areas.
The proposed model also generalizes well to \textbf{ISIC 2019} external validation, outperforming the ResNet‑50 baseline. These findings demonstrate that the proposed approach provides a more effective and efficient solution for computer-aided skin cancer diagnosis. The generation codes, weights and confusion matrices are open sourced in \href{https://github.com/Grapesea/ASFF-ResNet50-enhanced}{\textbf{https://github.com/Grapesea/ASFF-ResNet50-enhanced}}. 

\end{abstract}
\section{Introduction}
\label{sec:intro}

Skin cancer is a major global public health concern with rising incidence rates~\cite{rf1}. According to recent statistics, approximately 108,420 new cases of skin cancer were diagnosed in the United States in 2020~\cite{Siegel2020ColorectalCS}, with projections estimating 97,610 new cases and 7,990 deaths in 2022~\cite{ijms23073478}. Early diagnosis is critical for patient survival; timely excision can be curative~\cite{rf2}, whereas delayed diagnosis significantly increases mortality risk~\cite{Balch2009FinalVO}. Currently, primary recognition relies heavily on visual inspection by dermatologists, a process subject to inter-observer variability and heavily dependent on clinician experience, leading to potential inconsistencies and reduced diagnostic efficiency.

To address these limitations, computer-aided diagnosis (CAD) systems have emerged as a valuable tool, despite inherent challenges such as uneven color distribution, blurred lesion boundaries, and inter-class similarity~\cite{Maiti2020ComputeraidedDO,Li2024ResidualCS}. With the advancement of Artificial Intelligence (AI), deep learning—particularly Convolutional Neural Networks (CNNs)—has become the dominant paradigm in medical image analysis~\cite{Dayananda2023AMCCNetAA,10.1007/978-981-96-9961-2_39}. Classic CNN architectures such as VGG-16, Inception, and ResNet have demonstrated superior performance compared to traditional manual methods. Among these, ResNet is particularly advantageous due to its shortcut connections, which mitigate the vanishing gradient problem and enable the training of deeper networks~\cite{rf4}. ResNet and its variants are therefore regarded as promising directions for skin lesion diagnosis~\cite{Tan2024SkinLR}. Furthermore, ResNet architectures offer favorable hardware efficiency: owing to their dense convolutions and compact activation maps, they exhibit a high compute-to-memory ratio that is well-suited to modern accelerators. Empirically, ResNet-RS models have been reported to train up to 2.7× faster than EfficientNet-B6 on GPUs and up to 4.7× faster than EfficientNet-B5 on TPUs, yielding a superior speed–accuracy Pareto frontier~\cite{bello2021resnetrs}.

While more recent architectures such as EfficientNet and Vision Transformers (ViTs), as well as methods incorporating segmentation masks or clinical metadata, have been proposed, they present distinct limitations for practical deployment. ViTs typically require large-scale pretraining to converge and incur high computational latency due to their quadratic self-attention complexity~\cite{dosovitskiy2021vit}. Segmentation-informed methods demand expensive pixel-level annotations, while metadata-dependent approaches are constrained by the availability of auxiliary clinical information. In contrast, ResNet architectures offer a well-balanced trade-off among inference speed, hardware compatibility, and classification performance, making them a robust and practical backbone for developing efficient, pure-vision diagnostic systems.

Despite remarkable progress, CNN-based methods still face significant challenges in dermoscopic image analysis: high inter-class similarity with subtle intra-class variation, diverse lesion morphologies with irregular boundaries, and prevalent image artifacts such as hair, rulers, and uneven illumination~\cite{PMID:31366280}. Standard ResNet-50, while effective, lacks the adaptive mechanism required to balance high-level semantic representations with low-level surface details, often leading to misclassification in ambiguous cases. To address these issues, this work proposes an improved ResNet-50 enhanced with an Adaptive Spatial Feature Fusion (ASFF) mechanism. By dynamically integrating intermediate-level detail features with high-level semantic features via learnable spatial weights, ASFF improves feature discrimination and alleviates overfitting. \textbf{The main contributions of this paper are summarized as follows:}

\begin{itemize}
    \item We implement and benchmark eleven CNN baseline models (LeNet-5, AlexNet, VGG-16, ResNet-34, ResNet-50, ResNet-101, MobileNet, SqueezeNet, DenseNet, Xception, and GoogleNet) on the \textbf{ISIC 2020 dataset}. Extensive comparative experiments identify \textbf{ResNet-50} as the model offering the most balanced trade-off between accuracy and computational efficiency.
    
    \item We design an improved ResNet-50 model incorporating the ASFF mechanism, which adaptively fuses features from \verb|conv4_block6_out| and \verb|conv5_block3_out| via learned spatial weights. Comprehensive experiments verify its effectiveness, achieving an accuracy of \textbf{93.182\%—a 2.224\%} improvement over the standard ResNet-50—with superior performance across precision, recall, specificity, F1 score, and AUC metrics.

    \item We conduct ablation studies to validate the contribution of the adaptive weighting component, and provide qualitative evaluation via \textbf{Grad-CAM} visualizations, demonstrating that the proposed model focuses more accurately on lesion-relevant regions while suppressing background artifacts. Cross-dataset generalization on \textbf{ISIC 2019} further confirms the robustness of the proposed approach.
\end{itemize}

\section{Proposed Method}
\label{sec:Method}

\subsection{Adaptive Spatial Feature Fusion}
\label{derive}

\begin{figure}[htbp]
    \centering
\includegraphics[width=0.50\textwidth]{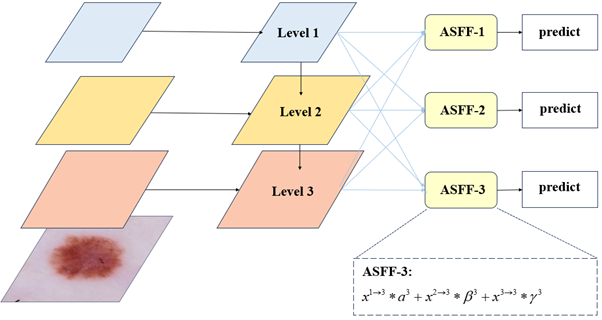}
    \caption{Structure of the proposed ASFF method.}
\label{fig:ASFF}
\end{figure}

Multi-scale feature fusion integrates semantic and low-level features across different scales, which is crucial for identifying small lesion regions susceptible to noise. This approach enhances classification by combining global context with local details. To improve efficiency beyond standard transfer learning or simple ensemble methods, we propose an Adaptive Spatial Feature Fusion (ASFF) method, as shown in the figure in Fig.~\ref{fig:ASFF}.

ASFF's key idea is to adaptively learn the spatial weight of fusion for feature maps at each scale. It consists of two steps: identically rescaling and adaptively fusing.

\paragraph{Feature Resizing.} We denote the features of the resolution at level $l$ ($l \in \{1,2,3\}$ for YOLOv3) as $\mathbf{x}^l$. For level $l$, we resize the features $\mathbf{x}^n$ at the other level $n$ ($n 
eq l$) to the same shape as that of $\mathbf{x}^l$. Because the features at three levels in YOLOv3 have different resolutions as well as different numbers of channels, we accordingly modify the up-sampling and down-sampling strategies for each scale. For up-sampling, we first apply a $1 \times 1$ convolution layer to compress the number of channels of features to that in level $l$, and then upscale the resolutions respectively with interpolation. For down-sampling with $1/2$ ratio, we simply use a $3 \times 3$ convolution layer with a stride of 2 to modify the number of channels and the resolution simultaneously. For the scale ratio of $1/4$, we add a 2-stride max pooling layer before the 2-stride convolution.

\paragraph{Adaptive Fusion.} Let $\mathbf{x}_{ij}^{n \rightarrow l}$ denote the feature vector at the position $(i, j)$ on the feature maps resized from level $n$ to level $l$. We propose to fuse the features at the corresponding level $l$ as follows:

\begin{equation}
    \mathbf{y}_{ij}^{l} = \alpha_{ij}^{l} \cdot \mathbf{x}_{ij}^{1 \rightarrow l}
    + \beta_{ij}^{l} \cdot \mathbf{x}_{ij}^{2 \rightarrow l}
    + \gamma_{ij}^{l} \cdot \mathbf{x}_{ij}^{3 \rightarrow l},
    \label{eq:asff1}
\end{equation}

where $\mathbf{y}_{ij}^l$ implies the $(i,j)$-th vector of the output feature maps $\mathbf{y}^l$ among channels. $\alpha_{ij}^l$, $\beta_{ij}^l$ and $\gamma_{ij}^l$ refer to the spatial importance weights for the feature maps at three different levels to level $l$, which are adaptively learned by the network. Note that $\alpha_{ij}^l$, $\beta_{ij}^l$ and $\gamma_{ij}^l$ can be simple scalar variables, which are shared across all the channels. We force $\alpha_{ij}^l + \beta_{ij}^l + \gamma_{ij}^l = 1$ and $\alpha_{ij}^l, \beta_{ij}^l, \gamma_{ij}^l \in [0,1]$, and define

\begin{equation}
    \alpha_{ij}^l = \frac{e^{\lambda_{\alpha_{ij}}^l}}{e^{\lambda_{\alpha_{ij}}^l} + e^{\lambda_{\beta_{ij}}^l} + e^{\lambda_{\gamma_{ij}}^l}}.
    \label{eq:asff2}
\end{equation}

Here $\alpha_{ij}^l$, $\beta_{ij}^l$ and $\gamma_{ij}^l$ are defined by using the softmax function with $\lambda_{\alpha_{ij}}^l$, $\lambda_{\beta_{ij}}^l$ and $\lambda_{\gamma_{ij}}^l$ as control parameters respectively. We use $1 \times 1$ convolution layers to compute the weight scalar maps $\lambda_\alpha^l$, $\lambda_\beta^l$ and $\lambda_\gamma^l$ from $\mathbf{x}^{1\rightarrow l}$, $\mathbf{x}^{2\rightarrow l}$ and $\mathbf{x}^{3\rightarrow l}$ respectively, and they can thus be learned through standard back-propagation.

With this method, the features at all the levels are adaptively aggregated at each scale. The outputs $\{\mathbf{y}^1, \mathbf{y}^2, \mathbf{y}^3\}$ are used for object detection following the same pipeline of YOLOv3.

\subsubsection{Consistency Property}

In this section, we analyze the consistency property of the proposed ASFF approach and the other alternatives of feature fusion. Without loss of generality, we focus on the gradient at a certain position $(i,j)$ of the unresized feature maps at level 1 $\mathbf{x}^1$ in YOLOv3. Following the chain rule, the gradient is computed as:

\begin{equation}
    \frac{\partial \mathcal{L}}{\partial \mathbf{x}_{ij}^1} =
    \frac{\partial \mathbf{y}_{ij}^1}{\partial \mathbf{x}_{ij}^1} \cdot \frac{\partial \mathcal{L}}{\partial \mathbf{y}_{ij}^1}
    + \frac{\partial \mathbf{x}_{ij}^{1\rightarrow 2}}{\partial \mathbf{x}_{ij}^1} \cdot \frac{\partial \mathbf{y}_{ij}^2}{\partial \mathbf{x}_{ij}^{1\rightarrow 2}} \cdot \frac{\partial \mathcal{L}}{\partial \mathbf{y}_{ij}^2}
    + \frac{\partial \mathbf{x}_{ij}^{1\rightarrow 3}}{\partial \mathbf{x}_{ij}^1} \cdot \frac{\partial \mathbf{y}_{ij}^3}{\partial \mathbf{x}_{ij}^{1\rightarrow 3}} \cdot \frac{\partial \mathcal{L}}{\partial \mathbf{y}_{ij}^3}
    \label{eq:grad1}
\end{equation}

It is worth to note that feature resizing usually uses interpolation for up-sampling and pooling for down-sampling. We thus assume that $\frac{\partial \mathbf{x}_{ij}^{1\rightarrow l}}{\partial \mathbf{x}_{ij}^1} \approx 1$ for simplicity. Then Eq.~(\ref{eq:grad1}) can be written as:

\begin{equation}
    \frac{\partial \mathcal{L}}{\partial \mathbf{x}_{ij}^1} =
    \frac{\partial \mathbf{y}_{ij}^1}{\partial \mathbf{x}_{ij}^1} \cdot \frac{\partial \mathcal{L}}{\partial \mathbf{y}_{ij}^1}
    + \frac{\partial \mathbf{y}_{ij}^2}{\partial \mathbf{x}_{ij}^{1\rightarrow 2}} \cdot \frac{\partial \mathcal{L}}{\partial \mathbf{y}_{ij}^2}
    + \frac{\partial \mathbf{y}_{ij}^3}{\partial \mathbf{x}_{ij}^{1\rightarrow 3}} \cdot \frac{\partial \mathcal{L}}{\partial \mathbf{y}_{ij}^3}
    \label{eq:grad2}
\end{equation}

For the two common fusion operations used in RetinaNet~\cite{ref21}, YOLOv3~\cite{ref31} and other pyramidal feature based detectors (\textit{i.e.} element-wise sum and concatenation), we can further simplify the equation to the following with $\frac{\partial \mathbf{y}_{ij}^1}{\partial \mathbf{x}_{ij}^1} = 1$ and $\frac{\partial \mathbf{y}_{ij}^2}{\partial \mathbf{x}_{ij}^{1\rightarrow 2}} = 1$:

\begin{equation}
    \frac{\partial \mathcal{L}}{\partial \mathbf{x}_{ij}^1} =
    \frac{\partial \mathcal{L}}{\partial \mathbf{y}_{ij}^1}
    + \frac{\partial \mathcal{L}}{\partial \mathbf{y}_{ij}^2}
    + \frac{\partial \mathcal{L}}{\partial \mathbf{y}_{ij}^3}
    \label{eq:grad3}
\end{equation}

Suppose position $(i,j)$ at level 1 is assigned as the center of an object according to a certain scale matching mechanism and $\frac{\partial \mathcal{L}}{\partial \mathbf{y}_{ij}^1}$ is the gradient from the positive sample. As the corresponding positions are viewed as background in the other levels, $\frac{\partial \mathcal{L}}{\partial \mathbf{y}_{ij}^2}$ and $\frac{\partial \mathcal{L}}{\partial \mathbf{y}_{ij}^3}$ are the gradients from negative samples. This inconsistency disturbs the gradient of $\frac{\partial \mathcal{L}}{\partial \mathbf{x}_{ij}^1}$ and downgrades the training efficiency of the original feature maps $\mathbf{x}^1$.

One typical way to deal with this problem is to set the corresponding positions of the other levels as ignore regions (\textit{i.e.} $\frac{\partial \mathcal{L}}{\partial \mathbf{y}_{ij}^2} = \frac{\partial \mathcal{L}}{\partial \mathbf{y}_{ij}^3} = 0$). However, although the conflict in $\mathbf{x}_{ij}^1$ is eliminated, the relaxation in $\mathbf{y}_{ij}^2$ and $\mathbf{y}_{ij}^3$ tends to cause more inferior predictions as false positives at the suboptimal levels.

For ASFF, it is straightforward to calculate the gradient from Eq.~(\ref{eq:asff1}) and Eq.~(\ref{eq:grad2}) as follows:

\begin{equation}
    \frac{\partial \mathcal{L}}{\partial \mathbf{x}_{ij}^1} =
    \alpha_{ij}^1 \cdot \frac{\partial \mathcal{L}}{\partial \mathbf{y}_{ij}^1}
    + \alpha_{ij}^2 \cdot \frac{\partial \mathcal{L}}{\partial \mathbf{y}_{ij}^2}
    + \alpha_{ij}^3 \cdot \frac{\partial \mathcal{L}}{\partial \mathbf{y}_{ij}^3},
    \label{eq:grad4}
\end{equation}

where $\alpha_{ij}^1, \alpha_{ij}^2, \alpha_{ij}^3 \in [0,1]$. With these three coefficients, the inconsistency of gradient can be harmonized if $\alpha_{ij}^2 \rightarrow 0$ and $\alpha_{ij}^3 \rightarrow 0$. Since the fusion parameters can be learned by the standard back-propagation algorithm, a well-tuned training process can yield such effective coefficients. Meanwhile, the supervision information of the background in $\frac{\partial \mathcal{L}}{\partial \mathbf{y}_{ij}^2}$ and $\frac{\partial \mathcal{L}}{\partial \mathbf{y}_{ij}^3}$ is kept, avoiding generating more false positives.

\subsection{ASFF Based ResNet-50}

\begin{figure}[htbp]
    \centering
    
    \includegraphics[width=0.75\textwidth]{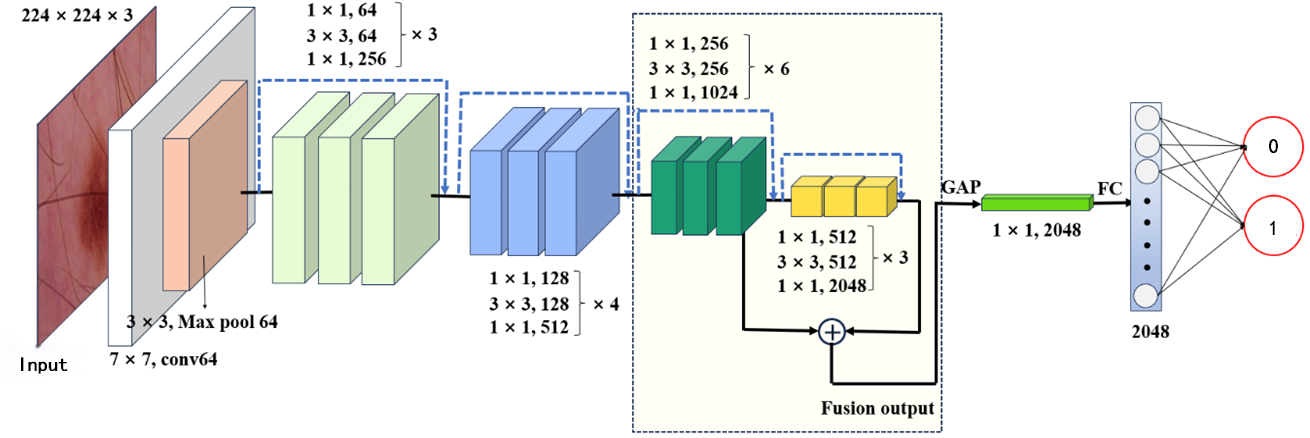}
    \caption{Structure of the proposed ResNet-50 based on ASFF.}
    \label{fig:ASFF2}

    \vspace{2em}
    
    \includegraphics[width=0.75\textwidth]{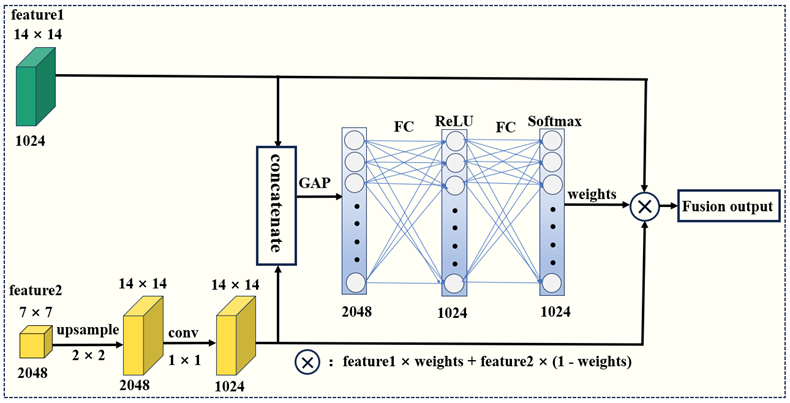}
    \caption{Detailed process of ASFF.}
    \label{fig:ASFF3}
\end{figure}

To improve the binary classification of skin lesion images, we integrated the ASFF mechanism into the ResNet-50 architecture. While the standard ResNet-50 fuses features hierarchically, it lacks the adaptivity required to balance semantic content and surface details, often leading to misclassification in complex cases. Our ASFF-based ResNet-50 introduces adaptive multi-scale weighting to emphasize lesion-relevant information and suppress redundant features, as illustrated in Fig.~\ref{fig:ASFF2}.

The ASFF process within our model is detailed in Fig.~\ref{fig:ASFF3}. In ResNet-50, \verb|conv4_block6_out| provides detailed feature maps, while \verb|conv5_block3_out| offers more abstract, semantic features. To align them, the output of \verb|conv5_block3_out| is upsampled to $14 \times 14$ and convolved to match the channel dimension of \verb|conv4_block6_out|. After concatenation, Global Average Pooling (GAP) and a fully connected layer are applied to obtain the fusion weights. The fusion result is formulated as Eq.~\ref{eq:asffresnet1}:

\begin{equation}
\text{Fusion output} = F_1 \times \omega + F_2 \times (1 - \omega)
\label{eq:asffresnet1}
\end{equation}

Here, $F_1$ and $F_2$ denote the outputs of \verb|conv4_block6_out| and \verb|conv5_block3_out|, respectively, and $\omega$ is the adaptive weight updated during training. This method efficiently integrates contextual information while maintaining a lightweight architecture.

The implementation of the proposed model proceeds in three stages. First, we extract feature maps from \verb|conv4_block6_out| ($14 \times 14$) and \verb|conv5_block3_out| ($7 \times 7$). Due to resolution differences, the deeper feature map is upsampled to match the spatial dimensions of the shallower map. Second, the features are concatenated and passed through a GAP layer followed by two fully connected layers: the first reduces dimensionality using ReLU activation, while the second applies Softmax to generate the adaptive weights. Finally, the fused features are processed by the classification head to output the binary prediction. The detailed pseudocode is implemented below.

\begin{figure}
    \centering
    \includegraphics[width=\linewidth]{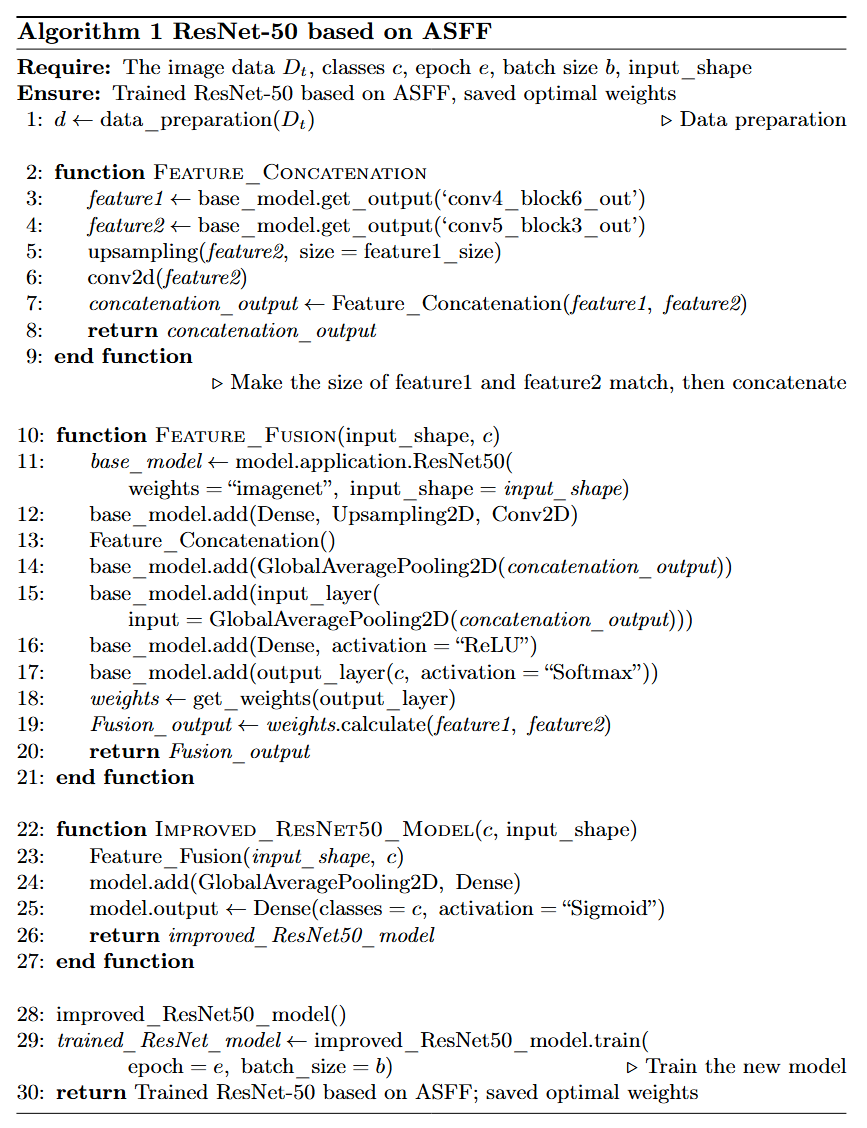}
    \label{fig:ASFF-ResNet50 Pseudocode}
\end{figure}

\section{Experimental Evaluation}
\label{sec:Experiment}

\subsection{Implementation Details}

Experiments were conducted on a workstation with an Intel Core i9-9900K CPU and an NVIDIA GeForce RTX 2080 Ti GPU (11GB), using Python 3.10.13 and TensorFlow 2.9.0. All models were trained for 500 epochs; hyperparameter configurations are detailed in Table~\ref{tab:models}.
The present study focuses on evaluating the proposed ASFF module within a controlled CNN.based framework rather than conducting an exhaustive benchmark over all recent backbone families. Therefore, we selected representative CNN architectures as baselines and used ResNet-50 as the main backbone for module-level improvement and ablation analysis. Recent architectures such as EfficientNet, ConvNeXt, and Vision Transformers are not included in the final quantitative comparison because they require substantially different model-scaling strategies, optimization recipes, input-resolution settings, and pretraining protocols. Including them without a dedicated and independently tuned benchmark may confound the evaluation of the proposed ASFF mechanism. We therefore regard comprehensive comparison with these recent architectures as future work~\cite{dosovitskiy2021vit}.

\begin{table}[htbp]
\caption{Experimental models and hyperparameters.}
  \centering
  \begin{tabular}{lcccccc}
  \hline
  \textbf{Model} & LeNet-5 & VGG-16 & ResNet-34 & ResNet-101 & ResNet-50 & MobileNet \\ \hline
  \textbf{LR} & 0.0001 & 0.0001 & 0.0005 & 0.0001 & 0.0005 & 0.0001  \\
  \textbf{Optimizer} & SGD & SGD & Adam & Adam & SGD & Adam \\
  \hline
  \textbf{Model} &
  SqueezeNet & DenseNet & AlexNet & Xception & GoogleNet & ASFF ResNet-50 \\ \hline
  \textbf{LR} & 0.0001 & 0.0001 &
  0.0001 & 0.0001 & 0.0001 & 0.0001 \\
  \textbf{Optimizer} & Adam & Adam & Adam & Adam &
  Adam & Adam \\
  \hline
  \end{tabular}
  \label{tab:models}
\end{table}

\subsection{Comparison with Prior Works}

To contextualize our results, we compare the proposed ASFF-based ResNet-50 against representative prior works evaluated on the ISIC 2020 dataset. Sayed~et~al.~\cite{sayed2021squeezenet} applied an optimized SqueezeNet with bald eagle search optimization to address class imbalance in ISIC 2020, reporting 98.37\% accuracy and 99\% AUC under a heavily oversampled setting. Cassidy~et~al.~\cite{cassidy2022isic} conducted a rigorous benchmark on ISIC 2020 following duplicate removal and class balancing, reporting a best AUC of only 0.80 across 19 models, underscoring the inherent difficulty of this dataset under controlled experimental conditions. Our ASFF-based ResNet-50 achieves an AUC of 0.9717 on a similarly balanced subset, demonstrating substantial improvement in discriminative ability. It is noteworthy that methods reporting higher accuracy typically rely on heavy oversampling, large-scale external pretraining, or multi-modal inputs, whereas our approach adopts a pure-vision pipeline trained on a moderately balanced subset of 3,297 images, making direct numerical comparison limited but our results competitive within this constrained setting.

\subsection{Data Description}

All experiments were conducted on a subset of the ISIC 2020 Challenge dataset, a public dermoscopic image collection with pathology‑confirmed labels; the ISIC 2019 Challenge dataset was used as an external validation set. The original dataset exhibits a severe class imbalance, which can bias CNN models toward the majority class. To mitigate this issue and ensure a rigorous evaluation of the model's discriminative ability, we constructed a balanced dataset comprising 3,297 images (1,800 benign and 1,497 malignant, using random seed generation to sample and split image datasets), and split into training and testing sets at a 4:1 ratio. This dataset presents significant challenges due to large intra-class variability, strong inter-class similarity, and diverse artifacts such as hair, rulers, and lighting variations, all of which heighten the risk of overfitting.

\subsection{Quantitative Results}

\begin{figure}[htbp]
\centering
\includegraphics[width = \textwidth]{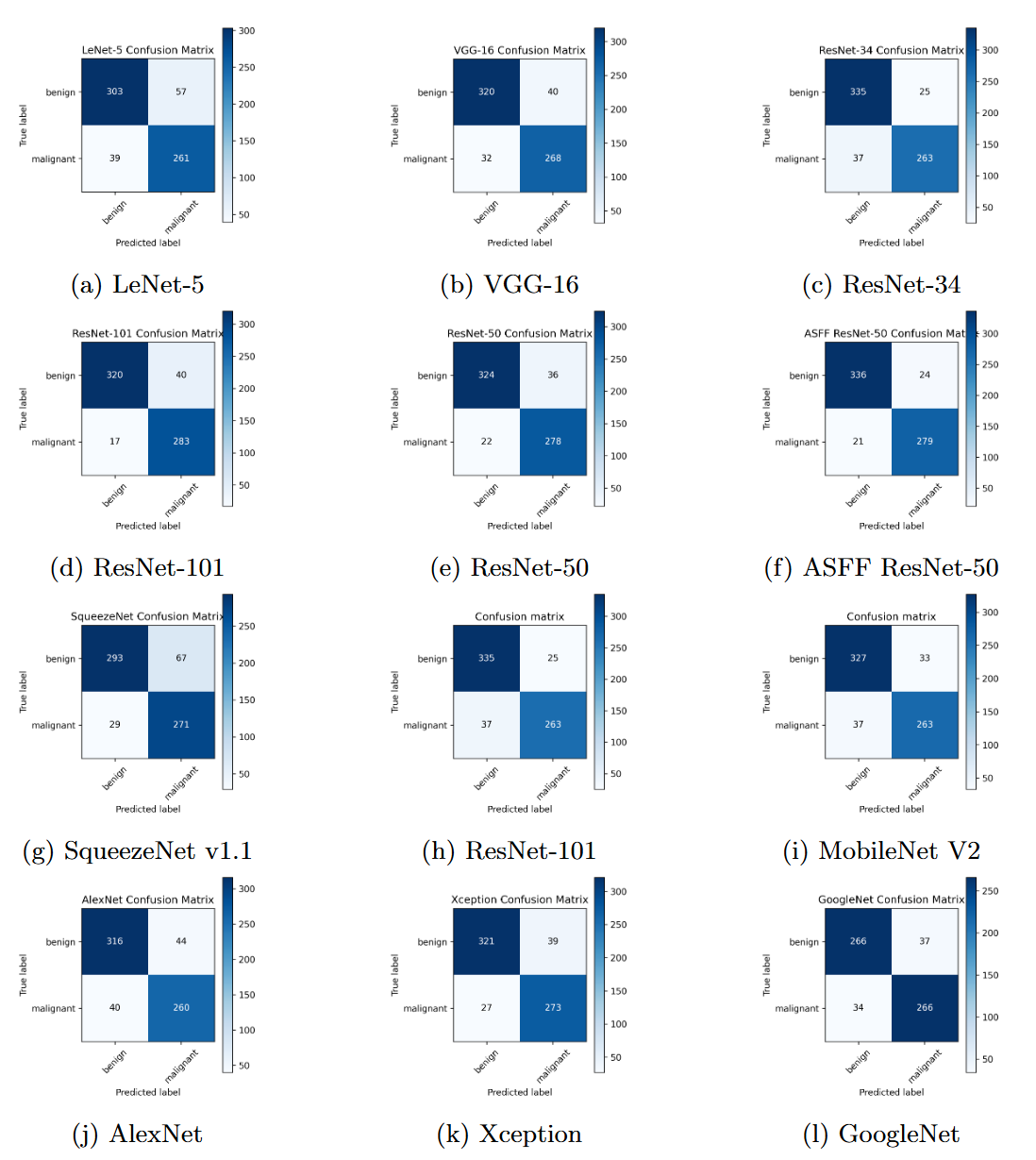}
\caption{Confusion matrices for the validated models.}
\label{fig:matrix}
\end{figure}

\begin{figure}[htbp]
    \centering
    \includegraphics[width=0.6\textwidth]{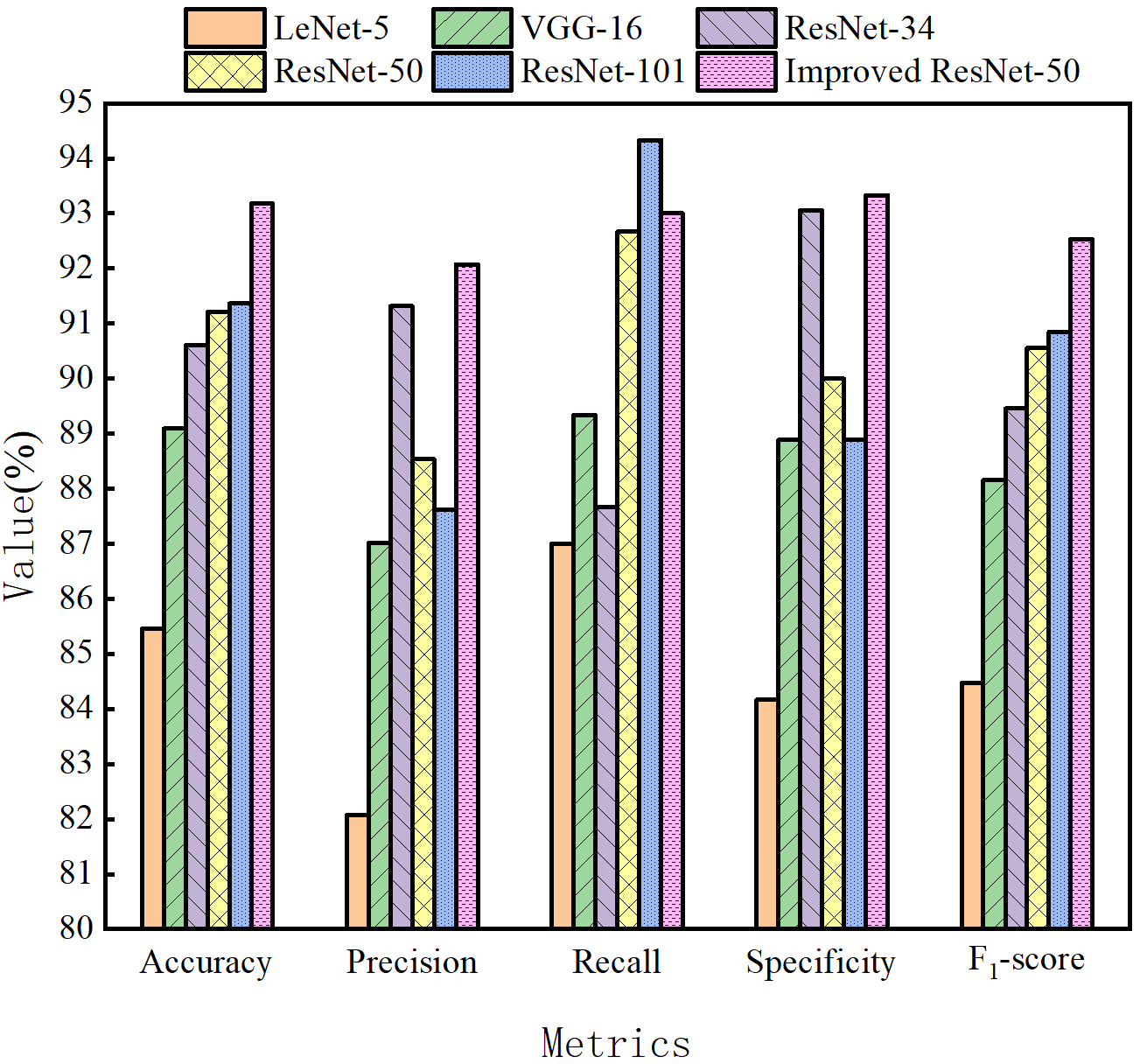}
    \caption{Performance of classification results.}
    \label{fig:datashow}

    \vspace{2em}
    
    \centering
    \begin{subfigure}[t]{0.48\linewidth}
        \centering
        \includegraphics[width=\linewidth]{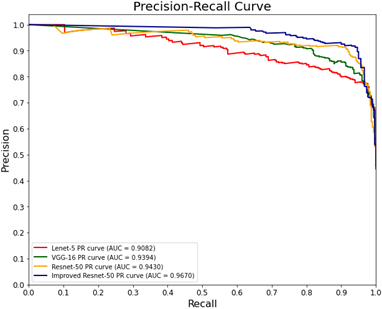}
        \caption{P-R Curve}
        \label{fig:pr}
    \end{subfigure}
    \hfill
    \begin{subfigure}[t]{0.48\linewidth}
        \centering
        \includegraphics[width=\linewidth]{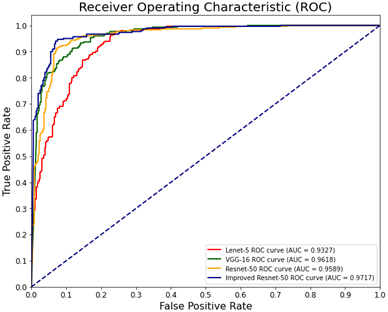}
        \caption{ROC Curve}
        \label{fig:roc}
    \end{subfigure}
    \caption{P-R and ROC curves for various models.}
    \label{fig:pr_roc}
\end{figure}

As illustrated in figures in Fig.\ref{fig:matrix}, most models achieve higher accuracy on benign samples, likely due to their simpler features and larger sample size. Among the selected CNN baselines under the same experimental protocol, the proposed ASFF-based ResNet-50 demonstrates the best overall performance, surpassing LeNet-5, VGG-16, ResNet-34, ResNet-101, AlexNet, SqueezeNet, Xception, GoogleNet and DenseNet. This improvement is attributed to more robust feature learning and reduced overfitting. Fig.~\ref{fig:datashow} and Table~\ref{tab:model_performance} further confirm its superiority; the proposed model achieves the highest scores across accuracy, precision, recall, specificity, and F1 score, reaching an overall accuracy of 93.182\%.

To verify the effectiveness of the proposed Adaptive Spatial Feature Fusion (ASFF) mechanism and investigate the contribution of different components, we conducted ablation studies on the ISIC 2020 dataset.We analyzed the impact of the adaptive weighting mechanism compared to rigid feature fusion methods and designed four variants:
  
\begin{itemize}
    \item \textbf{Baseline (ResNet-50):} The original network without any feature fusion.
    \item \textbf{Direct Summation:} Features from \verb|conv4_block6_out| and \verb|conv5_block3_out| are resized and added directly (element-wise sum) without learnable weights.
    \item \textbf{Direct Concatenation:} Features are concatenated along the channel dimension without adaptive re-weighting.
    \item \textbf{Proposed (ASFF):} The proposed method with learnable adaptive spatial weights.
\end{itemize}

The results are presented in Table~\ref{tab:ablation_fusion}. Direct concatenation gives a small improvement over the baseline in accuracy and specificity, while direct summation does not surpass the baseline. The proposed ASFF achieves the highest accuracy and F1 score overall.

To evaluate cross‑dataset generalization, we directly tested models trained on ISIC2020 on the ISIC2019 dataset without any fine‑tuning. The result shows that ResNet50-enhanced ASFF outperforms baseline. The external validation set is selected randomly from ISIC 2019 Challenge dataset and contains 1800 benign and 1497 malignant images. Table~\ref{tab:external} reports the results, where ASFF‑ResNet50 consistently outperforms the ResNet‑50 baseline, indicating stronger generalization.

\begin{table}
\caption{Model performance evaluation metrics.}
\centering
\begin{tabular}{lccccc}
\hline
\textbf{Model} & \textbf{Accuracy} & \textbf{Precision} & \textbf{Recall} & \textbf{Specificity} & \textbf{F1 Score} \\ \hline
SqueezeNet v1.1 & 85.455 & 80.178 & 90.333 & 81.389 & 84.953 \\
LeNet-5 & 85.455 & 85.336 & 85.583 & 78.333 & 85.395 \\
AlexNet & 87.278 & 85.526 & 86.667 & 87.779 & 86.093 \\
Xception & 88.030 & 83.587 & 91.667 & 85.000 & 87.440 \\
VGG-16 & 89.091 & 88.961 & 89.111 & 81.667 & 89.023 \\
GoogleNet & 89.242 & 87.789 & 88.667 & 89.722 & 88.226 \\
MobileNetV2 & 89.394 & 88.851 & 87.667& 90.833 & 88.255\\
DenseNet121 & 90.606 & 91.319 & 87.667 & 93.056 & 89.456\\
ResNet-34 & 90.606 & 90.687 & 90.361 & 83.056 & 90.493 \\
ResNet-101 & 91.364 & 91.286 & 91.611 & 83.750 & 91.336 \\
\textbf{ResNet-50} & \textbf{90.958} & \textbf{90.742} & \textbf{90.333} & \textbf{83.611} & \textbf{90.769} \\
\hline
\end{tabular}
\label{tab:model_performance}
\end{table}

\begin{table}[htbp]
\caption{Ablation study on different fusion strategies.}
    \centering
    \begin{tabular}{lccccc}
    \hline
      \textbf{Method} & \textbf{Accuracy} & \textbf{Precision} & \textbf{Recall} &
      \textbf{Specificity} & \textbf{F1 Score} \\ \hline
      Baseline (ResNet-50) & 90.958 & 90.742 & 90.333 & 83.611 & 90.769 \\
      ResNet-50 + Direct Sum & 91.061 & 90.745 & 90.667 & 83.674 & 91.169 \\
      ResNet-50 + Concat & 91.212 & 91.866 & 91.742 & 84.022 & 91.850 \\
      \textbf{Proposed (ASFF)} & \textbf{93.182} & \textbf{93.098} & \textbf{93.161} &
      \textbf{85.417} & \textbf{93.131} \\ \hline
      \end{tabular}
      \label{tab:ablation_fusion}
\end{table}

\begin{table}[htbp]
\caption{External validation on ISIC2019.}
    \centering
    \begin{tabular}{lccccc}
    \hline
    \textbf{Model} & \textbf{Accuracy} & \textbf{Precision} & \textbf{Recall} &
    \textbf{Specificity} & \textbf{F1 Score} \\ \hline
    ResNet-50 & 0.7652 & 0.7305 & 0.7652 & 0.7652 & 0.7474 \\
    ASFF ResNet-50 & 0.8015 & 0.7705 & 0.8015 & 0.8015 & 0.7857 \\
    \hline
    \end{tabular}
    \label{tab:external}
\end{table} 

Fig.\ref{fig:pr_roc}a indicates that LeNet-5 performs poorly, with precision dropping rapidly as sensitivity increases, reflecting its limited capacity for complex dermoscopic data. ResNet-50 generally outperforms LeNet-5 and VGG-16; however, in the sensitivity range of 0.4–0.6, its performance dips below that of VGG-16, suggesting potential limitations in extracting certain representative features. Nevertheless, ResNet-50 largely benefits from its deeper residual connections. The improved ResNet-50 surpasses all other models across the entire curve, validating the effectiveness of adaptive feature fusion. Similarly, in Fig.\ref{fig:pr_roc}b, LeNet-5 exhibits the lowest AUC. While VGG-16 slightly exceeds the standard ResNet-50 at lower thresholds, the improved ResNet-50 achieves the best result with an AUC of 0.9717. Its curve remains consistently above the others, demonstrating that adaptive feature fusion effectively balances surface and semantic features, optimizing recognition for both simple and complex samples.

\subsection{Visualization based on Grad-CAM}

\begin{figure}
\centering
    \includegraphics[width=0.8\textwidth]{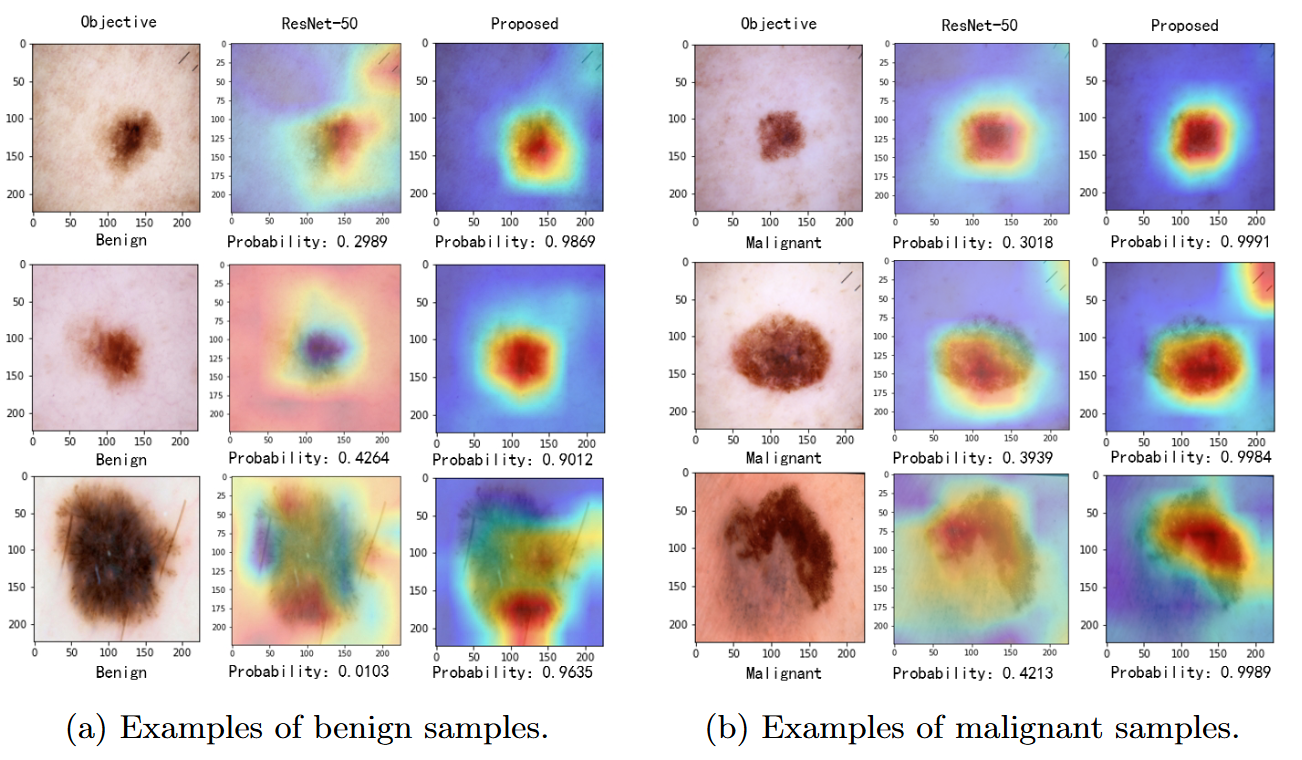}
\caption{Examples of samples correctly predicted only by the proposed model (Grad-CAM visualization).}
\label{fig:b2_m2}
\end{figure}

Fig.\ref{fig:b2_m2} presents qualitative Grad-CAM results. For benign samples (Fig.\ref{fig:b2_m2}a), which are complicated by visual similarity to malignant lesions, color overlap, and noise, ResNet-50 misclassifies them with heatmaps focused on irrelevant artifacts such as hair and background, whereas the improved model accurately localizes the lesion area. For malignant samples (Fig.\ref{fig:b2_m2}b), ResNet-50 roughly locates the lesion but fails to capture fine-grained features, while the improved model highlights lesion regions with greater precision. Overall, the ASFF-based network demonstrates superior robustness to noise, particularly under low contrast between the lesion and surrounding skin.

\section{Discussion and Conclusion}
\label{sec:conclusion}

This study addressed the challenges of dermoscopic image classification, including high inter-class similarity, intra-class variability, and prevalent image artifacts. To establish a comprehensive baseline, we systematically benchmarked eleven CNN architectures: LeNet-5, AlexNet, VGG-16, ResNet-34, ResNet-50, ResNet-101, MobileNetV2, SqueezeNet v1.1, DenseNet121, Xception, and GoogleNet on a balanced subset of the ISIC 2020 dataset, identifying ResNet-50 as the most favorable backbone in terms of accuracy and computational efficiency.

To overcome ResNet-50's limitations in multi-scale feature extraction, we proposed an Adaptive Spatial Feature Fusion (ASFF) framework that dynamically integrates intermediate and high-level features via learnable spatial weights, enabling the model to emphasize lesion-relevant regions while suppressing background artifacts. The ASFF-based ResNet-50 achieves 93.182\% accuracy (+2.224\% over baseline), with P-R AUC of 0.9670 and ROC AUC of 0.9717. Ablation studies confirm that adaptive weighting is the key driver of improvement over rigid fusion alternatives. Cross-dataset evaluation on ISIC 2019 without fine-tuning further yields 80.15\% accuracy (+3.63\% over ResNet-50), demonstrating stronger generalization. Grad-CAM visualizations qualitatively confirm more accurate lesion localization compared to the standard baseline.

Despite these promising results, the current model is trained on a moderately sized subset and fuses only two feature levels, leaving room for improvement. Future work will explore hierarchical multi-level fusion, integration of clinical metadata for multimodal classification, and lightweight variants of ASFF for deployment on resource-constrained clinical devices.

\begin{acknowledgments}
Support was provided by National Key Research and Development Program of China (Grant No. 2024YFC2511003). The authors declare that there are no conflicts of interest or competing financial interests related to this study.
\end{acknowledgments}

%
%
%
%

\end{document}